\documentclass{article}
\usepackage[utf8]{inputenc}
\usepackage{amsmath}
\usepackage{amsfonts}
\usepackage{dsfont} 
\usepackage{amssymb} 
\usepackage{ragged2e}
\usepackage{bbm} 
\usepackage{stmaryrd} 
\usepackage{placeins} 
\usepackage{diagbox}

\usepackage[nottoc]{tocbibind}

\usepackage{titling}
\newcommand{\subtitle}[1]{%
  \posttitle{%
    \par\end{center}
    \begin{center}\large#1\end{center}
    \vskip0.5em}%
}

\usepackage{graphicx} 

\usepackage{float}

\usepackage{titling}
\usepackage{hyperref}

\usepackage{multirow}
\usepackage{makecell}
\usepackage{adjustbox} 

\usepackage{enumitem}

\usepackage[utf8]{inputenc} 
\usepackage[T1]{fontenc}    
\usepackage{textcomp}       

\usepackage[utf8]{inputenc}
\usepackage{algorithm}
\usepackage[noend]{algpseudocode}

\usepackage[a4paper, left=2.54cm, right=2.54cm, top=2.54cm, bottom=2cm]{geometry}

\usepackage[english]{babel}

\usepackage{fancyhdr}

\usepackage{amsthm}

\theoremstyle{definition}

\begin{document}

\title{SigBERT: Combining Narrative Medical Reports and Rough Path Signature Theory for Survival Risk Estimation in Oncology}

\author{
    Paul MINCHELLA\textsuperscript{1,2,3}, Loïc VERLINGUE\textsuperscript{2}, St\'ephane CHRETIEN\textsuperscript{3}, \\ R\'emi VAUCHER\textsuperscript{3}, Guillaume METZLER\textsuperscript{3} \\
    \\
    \small{\textsuperscript{1}\textit{SHAPE-Med@Lyon, 74 Boulevard du 11 novembre 1918, 69100 Villeurbanne}}, \\ \small{\texttt{minchellapaul@gmail.com}}\\
    \small{\textsuperscript{2}\textit{Centre de Recherche en Canc\'erologie de Lyon, 19 Bd Jean XXIII, 69008 Lyon} }, \\ \small{\texttt{loic.verlingue@lyon.unicancer.fr}}\\
    \small{\textsuperscript{3}\textit{Laboratoire ERIC, 5 avenue Pierre Mendes-France, 69676 Bron}}, \\ \small{\texttt{\{stephane.chretien, r.vaucher, guillaume.metzler\}.univ-lyon2.fr}}
}

\date{\today}

\maketitle




\bigskip

\begin{abstract}
Electronic medical reports (EHR) contain a vast amount of information that can be leveraged for machine learning applications in healthcare. However, existing survival analysis methods often struggle to effectively handle the complexity of textual data, particularly in its sequential form. Here, we propose SigBERT, an innovative temporal survival analysis framework designed to efficiently process a large number of clinical reports per patient. SigBERT processes timestamped medical reports by extracting and averaging word embeddings into sentence embeddings. To capture temporal dynamics from the time series of sentence embedding coordinates, we apply signature extraction from rough path theory to derive geometric features for each patient, which significantly enhance survival model performance by capturing complex temporal dynamics. These features are then integrated into a LASSO-penalized Cox model to estimate patient-specific risk scores. The model was trained and evaluated on a real-world oncology dataset from the L\'eon B\'erard Center corpus, with a C-index score of $0.75 \ (\mathrm{sd}\ 0.014)$ on the independent test cohort. 
SigBERT integrates sequential medical data to enhance risk estimation, advancing narrative-based survival analysis.

\end{abstract}


\newpage

\section{Introduction}

\subsection{Background}

Survival analysis plays a fundamental role in medicine (oncology, cardiology, nephrology, critical care, etc.) where predicting the prognosis of the patient is crucial to guide clinical decision-making. They help determine treatment strategies, assess the efficacy of therapeutic interventions, refine clinical trial eligibility criteria, aid in risk stratification and early intervention planning. The Cox Proportional Hazards model \cite{cox_og} has long been the gold standard since it was published in survival analysis due to its interpretability and effectiveness in identifying prognostic factors. One of its key advantages is its ability to effectively account for censoring, which arises when the event of interest (\textit{e.g.}, death, relapse, or disease progression) has not yet occurred for certain patients by the end of the study period. Even without an observed event, censored patients provide valuable information by contributing to likelihood, as their recorded survival time improves risk estimation despite incomplete event data. This feature makes the model particularly robust in real-world clinical settings, where missing or censored data are very common. Over the past decade, more recent advances in survival analysis have explored neural network-based models, which offer a powerful alternative by capturing complex, non-linear relationships within patient data \cite{Katzman_2018}, \cite{lee2018deephit}. 

However, a key limitation of many survival models is their reliance on static patient snapshots rather than dynamic, time-dependent data. Integrating structured (\textit{e.g.}, biomarkers, lab tests) and unstructured data (\textit{e.g.}, clinical narratives) is crucial but intricate. Addressing these challenges requires advanced NLP methods for processing unstructured clinical narratives, along with statistical techniques to enhance predictive accuracy and clinical applicability.

\subsection{Related works}

Recent advancements in survival analysis have introduced dynamic models capable of integrating time-dependent patient data. Dynamic-DeepHit \cite{dynamicdeephit2019} employs RNNs with attention mechanisms to process sequential biomarkers and treatments. 
Transformer-based approaches like BERTSurv \cite{bertsurv} leverage pretrained language models to extract survival-relevant features from unstructured clinical notes, enhancing survival prediction. 
Meanwhile, CoxSig \cite{CoxSig} incorporates signature transforms and controlled differential equations to model time-dependent features. 


\begin{table}[h]
\centering
\begin{adjustbox}{width=\textwidth,center}
\renewcommand{\arraystretch}{1.2}
\begin{tabular}{|
>{\raggedright\arraybackslash}m{3.2cm}|
>{\raggedright\arraybackslash}m{4.7cm}|
>{\raggedright\arraybackslash}m{4.1cm}|
m{4.1cm}|
m{1.6cm}|
m{1.6cm}|
m{1.9cm}|}
\hline
\textbf{Model Paper} & 
\textbf{Field \& Datasets} & 
\textbf{Model Architecture} & 
\textbf{Sequential Features} & 
\textbf{C-index} & 
\textbf{td-AUC} & 
\textbf{BS} \\
\hline \hline

\textbf{SigBERT (Ours)} & 
Oncology, narrative reports (L\'eon B\'erard) & 
OncoBERT + Signature + Cox LASSO & 
Yes, NLP with path signatures & 
0.75 & 0.80 & $\ll$0.25 \\

\hline
\textbf{MSK-CHORD} \cite{Jee2024} & 
Oncology, Real-world (MSK-CHORD) & 
Random Survival Forest (RSF) & 
No, features at fixed time point & 
$[0.58,0.83]$ & – & – \\

\hline
\textbf{CoxSig} \cite{CoxSig} & 
Maintenance, synthetic + real (NASA, Califrais) & 
Cox model + Signature transforms & 
Yes, time-series encoded with Signature & 
– & $[0.74,0.87]$ & $[0.09,0.15]$ \\

\hline
\textbf{BERTSurv} \cite{bertsurv} & 
ICU (MIMIC-III, not oncology) & 
Transformer (BERT) & 
Yes, from sequential clinical notes (NLP) & 
0.7 & – & – \\

\hline
\textbf{DySurv} \cite{dysurv2024} & 
ICU (MIMIC-III, eICU) & 
CVAE + LSTM & 
Yes, sequential EHR (structured) & 
$\approx$0.60 & – & included \\

\hline
\textbf{Survival Seq2Seq} \cite{survival_seq2seq} & 
General (MIMIC-IV + synthetic) & 
Seq2Seq (GRU-D + Attention) & 
Yes, hospital time series & 
– & $[0.84,0.91]$ & – \\

\hline
\textbf{Dynamic-DeepHit} \cite{dynamicdeephit2019} & 
Cystic Fibrosis (UK Registry) & 
Deep RNN + Temporal Attention & 
Yes, repeated biomarker vectors & 
$[0.94, 0.96]$ & td-AUC & – \\

\hline
\textbf{DeepSurv} \cite{Katzman_2018} & 
General + oncology (e.g., METABRIC) & 
DNN with Cox PH loss & 
No, static baseline covariates & 
$[0.61, 0.86]$ & – & – \\

\hline
\textbf{Landmark Endpoint} \cite{Devaux2022} & 
Liver disease (PBC), Aging (PAQUID) & 
Landmark (Cox, RSF, penalized) & 
Yes, repeated biomarker measures & 
– & $[0.73, 0.87]$ & $[0.076, 0.089]$ \\

\hline
\textbf{Penalized Reg. Calib.} \cite{Signorelli_2021} & 
Neuromuscular (DMD, MARK-MD) & 
Penalized Cox + Mixed Effects & 
Yes, blood biomarker sequences & 
$[0.7, 0.8]$ & $[0.73, 0.87]$ & – \\

\hline
\end{tabular}
\end{adjustbox}
\caption{Overview of state-of-the-art survival models across domains. All reported metrics are taken from the original publications; no external re-evaluation was performed on our dataset.}
\label{tab:SOTA_tab}
\end{table}

Table \ref{tab:SOTA_tab} provides a survey and summarizes the performance of several recent survival models across different medical domains and datasets, highlighting their architectural choices and whether they incorporate sequential information. Our model SigBERT compares favorably with these methods. Specifically, it achieves a C-index of 0.75, a mean td-AUC of 0.794, and an IBS below 0.25 over 10 years-values that align well with or exceed several state-of-the-art models, such as CoxSig or Penalized Regression Calibration. 
While some deep learning approaches like Dynamic-DeepHit or Survival Seq2Seq achieve higher td-AUC, these results are often obtained on synthetic or ICU-based datasets with structured biomarker sequences, which differ substantially from our real-world oncology setting involving complex, unstructured textual data. It is also important to note that strict comparison remains challenging due to differing data modalities and availability, as most datasets used in these works are not publicly accessible or shareable due to clinical data protection policies. In contrast, our study demonstrates strong results on a large-scale real oncology dataset, highlighting the practical relevance and robustness of our method in operational conditions.

\subsection{Our Contributions}

Our approach contributes to survival analysis by leveraging rich representations from NLP-based embeddings, combined with signature transforms to capture the temporal dynamics of patients' follow-up, thus demonstrating the impact of unstructured clinical narratives in oncology risk estimation. 

Additionally, our pipeline supports multi-data integration, enabling future incorporation of structured patient data, such as tumor stage, demographic factors, and tumor topography. Moreover, the use of signature transforms allows us to process a very large number of clinical reports per patient without incurring a prohibitive computational cost. Unlike traditional models that struggle with long sequences due to high memory and time complexity, our method efficiently encodes all available follow-up data for each patient. Please refer to our GitHub repository at \texttt{\url{https://github.com/MINCHELLA-Paul/SigBERT}} for access to the code. The notebook \verb|SigBERT_study.ipynb| provides numerous complementary results.


\section{Method}

\begin{figure}
  \centering
  \includegraphics[width=\linewidth]{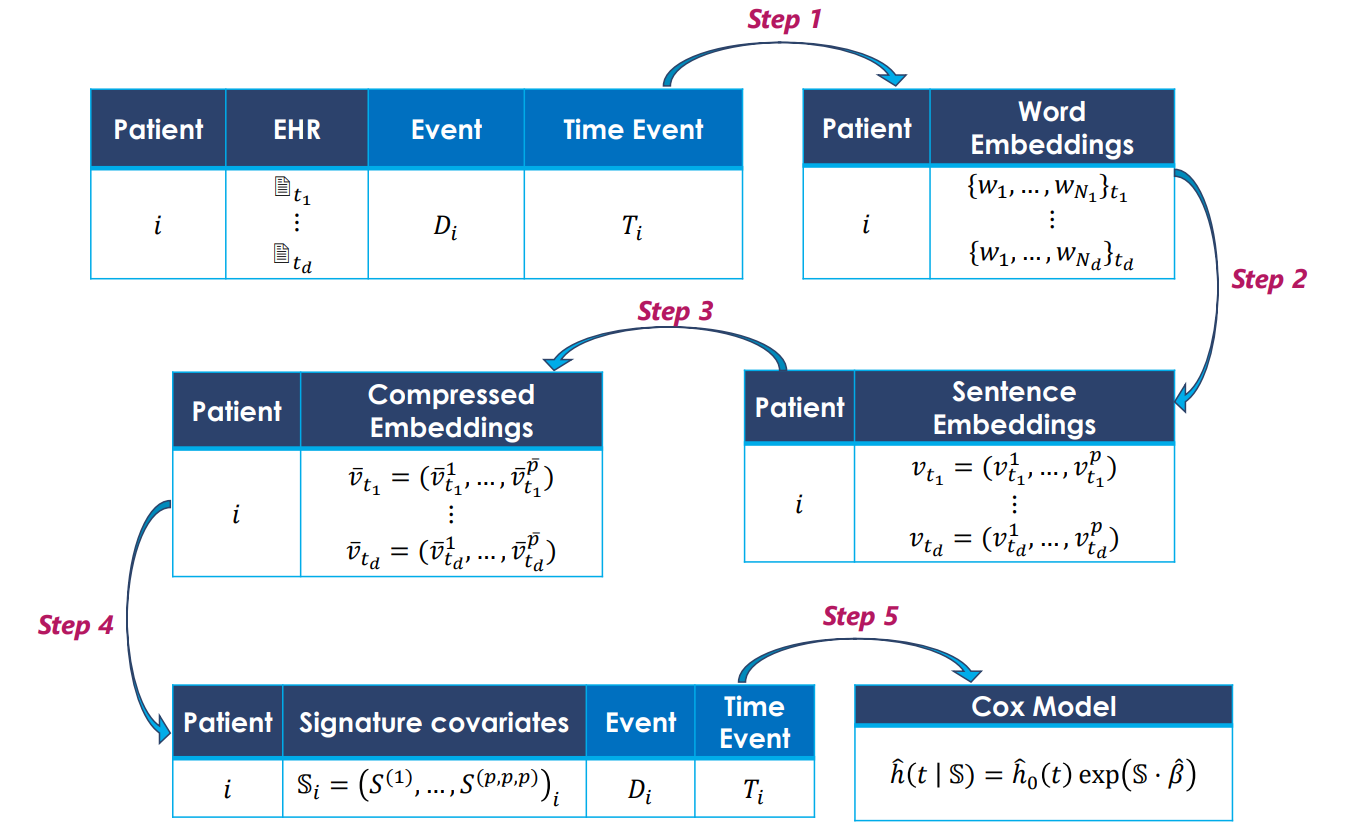}
  \caption{\textbf{Global Pipeline}: A five-step approach for SigBERT. \textbf{Step 1:} Extract word embeddings from medical reports using OncoBERT. \textbf{Step 2:} Compute sentence embeddings by averaging word embeddings for each report. \textbf{Step 3:} Compress the sentence embeddings through a dimensionality reduction mapping. \textbf{Step 4:} Apply the signature transform to extract coefficients capturing temporal dynamics as covariates. \textbf{Step 5:} Use a Cox model with LASSO regularization to estimate risk scores.}
  \label{fig:pipeline}
\end{figure}

Our dataset consists of the following information: Each patient is associated with a set of medical reports, each recorded at a specific timestamp, providing a longitudinal sequence of textual data. In addition, we have access to each patient's time in the study and event status, indicating whether the patient experienced the event of interest (\textit{e.g.}, death) or was censored. This forms the foundation for our survival analysis pipeline. One can refer to Figure \ref{fig:pipeline} at any time for an overview of our global pipeline.

Before computing embeddings, raw clinical reports are preprocessed to clean and standardize the textual data. The reports are loaded from structured files, with unnecessary columns removed. The text field (constituting the report at time $t$) is cleaned by stripping redundant metadata, such as the report source when it appears at the beginning. Duplicate reports are dropped to avoid repetition. In parallel, all date-related columns are converted to a consistent datetime format. This preprocessing step helps ensure the text input is clean and coherent, while preserving the temporal structure needed for downstream analysis.

To process long clinical texts, we kept the entire content of each medical report without truncation. Note that OncoBERT \cite{oncobert}, being a fine-tuned version of CamemBERT \cite{martin2020camembert}, inherits a maximum input length of 512 tokens due to its RoBERTa-based architecture. We will describe it in more detail in the following section. 
We did not apply any length filtering or summarization at this stage. 

Furthermore, to better simulate real-world clinical scenarios, we deliberately mask the last (at least) 100 observed days from each patient’s record. This ensures that predictions rely on earlier medical data rather than immediate pre-mortem indicators, improving generalization for prospective survival analysis.

\subsection{Embeddings computation}

The first step of our pipeline involves transforming unstructured medical reports into numerical representations using OncoBERT \cite{oncobert}. It is a CamemBERT-based language model fine-tuned on a large corpus of oncology-related clinical notes from L\'eon B\'erard Center, making it particularly suited for extracting meaningful representations from oncology-specific narratives. This approach enables us to numerically encode the semantic information embedded in the textual data while preserving the rich clinical context contained within patient histories. We obtain a dictionary mapping each word (token) to its corresponding embedding vector within the learned (fine-tuned) vector space of dimension $p=768$. It can be formally expressed as follows for a given patient $i$: \begin{equation*}
    \label{word_embd}
    \{ \text{word}_1 : v_{w_1} ,\ \dots, \ \text{word}_{N_t} : v_{w_{N_t}} \}_{i}
, \end{equation*} where $N_t$ is the number of unique tokens in the medical report at time $t$.

\newpage

It is worth emphasizing that the superior performance of OncoBERT is consistent with expectations for a risk estimation task in oncology based on French clinical narratives. OncoBERT is a domain-specific language model fine-tuned on cancer-related French medical reports, and thus better captures the linguistic and clinical nuances of the data (including specialized oncology vocabulary and domain-specific phrasing). As such, it is not surprising that this specialized NLP model outperforms more generic approaches: for instance, Word2Vec-based embeddings \cite{word2vec}, even with carefully tuned parameters, reach at most a concordance index of $0.6$, while CamemBERT-based embeddings reach at most $0.7$.

We aim to aggregate these representations into a single vector per report, so that each patient’s medical record at time $t$ is captured as a meaningful numerical representation. Thus, we employ the Smooth Inverse Frequency (SIF) method proposed by \cite{arora2017asimple}, a robust unsupervised approach for computing sentence embeddings. For a arbitrary report $s$, its representation is given by: \begin{equation*}
    \label{sentence_embd_arora}
    v_s = \frac{1}{|s|} \sum_{w \in s} \frac{a}{\mathrm{f}(w)+a} \ v_w \ \in \mathbb{R}^p
, \end{equation*} where $\mathrm{f}(w)$ represents the frequency of a given word in the corpus, and $a$ is a smoothing parameter (typically set to $10^{-3}$). 


While it is common practice to use the CLS token output from BERT-based models to represent entire input sequences, we explored an alternative approach for encoding clinical reports. Although the CLS token was initially considered in our pipeline, we ultimately adopted the SIF method based on retrospective evaluation results and its strong theoretical foundations. Specifically, SIF-based sentence embeddings yielded consistently better predictive performance, including a higher C-index (0.75 vs. 0.70), improved time-dependent AUC, and lower Brier Scores. We attribute these gains to the robustness of the SIF method, which re-weights and averages token-level embeddings to form a more stable and generalizable sentence-level representation. 

\newpage 

Moreover, SIF offers a practical advantage when handling long clinical texts: by aggregating token embeddings across the entire report - possibly by processing it in overlapping chunks - it allows us to capture information beyond the 512-token input limit inherent to BERT models. In contrast, the CLS token is extracted from a truncated version of the input (limited to 512 tokens) and its representation is tightly dependent on the pretraining and fine-tuning stages of the language model - whereas our OncoBERT model was not specifically optimized for the downstream use of CLS embeddings in survival prediction tasks.


At this point, a patient $i$ with $N_i$ clinical reports can be formally represented by a collection of $p$-dimensional time-indexed vectors as follows:
\begin{equation*}
    \label{patient_total_vectors}
    \left( {v}_{t_1}, \dots, {v}_{t_{N_i}} \right)^{\top}_i \in \mathbb{R}^{N_i \times p}.
\end{equation*}
We will define the signature transform in the following section, which is the mathematical framework used to extract features from time series for our survival model. For now, it is sufficient to note that for $p$ time series and a given truncation level $L$, the number of signature coefficients to compute is given by $\frac{p^{L+1} - 1}{p - 1} = O(p^L)$. This quantity grows exponentially with respect to the number of channels (\textit{i.e.}, time series dimensions) involved. For instance, with $p = 768$ and $L = 3$, this results in approximately $4.5 \times 10^8$ coefficients, which is computationally intractable.

We are thus compelled to reduce the dimensionality of the sentence embeddings. 
This requirement also brings significant advantages. The first is computational: reducing $p$ greatly improves the numerical feasibility of downstream processing. The second is theoretical: the original embedding space of dimension $p$ may be projected into a lower-dimensional space of size $\bar{p}$ while retaining most of the information relevant to our prediction task, namely risk estimation. By applying a linear transformation, we can map the embeddings into this compressed space and carry out computations much more efficiently.

A straightforward approach is to apply Principal Component Analysis (PCA) on all sentence embeddings in the training set to obtain a compression matrix $R_\text{comp}$. From a linear algebra perspective, $R_\text{comp} \in \mathbb{R}^{\bar{p} \times p}$ is a projection matrix whose rows correspond to the top $\bar{p}$ principal components, \textit{i.e.}, the orthonormal directions that capture the highest variance in the original $p$-dimensional embedding space. Applying $R_\text{comp}$ to each sentence embedding results in a lower-dimensional representation in $\mathbb{R}^{\bar{p}}$ that retains as much relevant information as possible in terms of variance. This baseline method is simple to implement, computationally efficient, and often yields satisfactory results in practice.

We define the compressed vector as $\bar{v}_s := R_\text{comp} \cdot v_s$. This transformation preserves most of the semantic information while reducing the dimensionality of the paths from $p=768$ to, for instance, $\bar{p}=25$. Then, a patient with $N_i$ reports can be formally represented by a family of $\bar{p}$ time series as follows: \begin{equation*}
    \label{patient_compressed_vectors}
    \left( \bar{v}_{t_1}, \dots, \bar{v}_{t_{N_i}}  \right)^{\top}_i \in \mathbb{R}^{ N_i \times \bar{p}}
. \end{equation*} 

The choice of this compression value is based on a retrospective study evaluating the trade-off between computational cost and the C-index achieved on the test set. By progressively increasing the compressed dimension ($\bar{p} = 10, 15, 20, 25, 30, \dots$), we observed convergence of the C-index on the test set starting from $\bar{p} = 25$. This value was identified as the optimal candidate, as it provides satisfactory performance while guaranteeing computational efficiency.

\subsection{Signature features extraction}

The signature of a path \cite{chen1954}, adapted for rough path theory by Terry Lyons  \cite{lyons1998differential}, provides a systematic method for encoding sequential data into geometric features, using iterated integrals. Consider a $p$-dimensional path (which means each coordinates form a path), denoted as $v = (v^1, \dots, v^p)$, defined over the interval $[0,T]$. For any integer $k \geq 1$, any sequence of indices $i_{1}, \ldots, i_{k} \in \{ 1, \dots, p\}$, any $0 \leq t_1 < \cdots < t_k \leq T $, the iterated integral signature of $v$ up to time $t \in [0,T]$ is defined as: \begin{equation*}
    \label{sign_def}
    S(v)_{0, t}^{(i_{1}, \ldots, i_{k})}=\int_{0<t_{k}<t} \cdots \int_{0<t_{1}<t_{2}} \mathrm{d} v_{t_{1}}^{i_{1}} \ldots \mathrm{d} v_{t_{k}}^{i_{k}}. 
\end{equation*} 

The collection of these features is organized in tensor form which uniquely encodes the path and is defined as: 
\begin{equation*}
    S^{k}(v) = \Big( S(v)_{0, T}^{(i_{1}, \ldots, i_{k})} \Big)_{(i_{1}, \ldots, i_{k}) \in \{ 1, \dots, p\}^k} \in (\mathbb{R}^p)^{\otimes k}.
\end{equation*}
Thus, the truncated signature up to order $L$ naturally belongs to the truncated tensor algebra $\mathcal{T}^{\leq L}(\mathbb{R}^p) = \bigoplus_{k=0}^{L} (\mathbb{R}^p)^{\otimes k}$ of order $L$ over $\mathbb{R}^p$ :
\begin{equation*}
    S^{\leq L}(v) = \big( S^{k}(v) \big)_{k=0}^{L} \in \mathcal{T}^{\leq L}(\mathbb{R}^p).
\end{equation*}

In addition to encoding temporal dynamics, this approach handles sequences of varying lengths and is invariant to translation and temporal reparameterization (see \cite{Chevyrev2016}), making it well-suited for patients with different study entry points and durations. Moreover, one of the most fundamental and computationally advantageous properties of the signature transform is Chen’s identity, which provides a recursive structure for computing signatures efficiently. Given two continuous paths $X: [a, b] \to \mathbb{R}^d$ and $Y: [b, c] \to \mathbb{R}^d$, their concatenation is defined as the path $X * Y: [a, c] \to \mathbb{R}^d$ such that:

$$
(X * Y)_t =
\begin{cases}
    X_t, & t \in [a, b], \\
    X_b + (Y_t - Y_b), & t \in [b, c].
\end{cases}
$$ Chen’s identity states that the signature of a concatenated path can be factorized as the tensor product of the signatures of its subpaths:

$$
S(X * Y) = S(X) \otimes S(Y).
$$ This property is particularly useful in computational applications, as it allows for the efficient computation of path signatures by processing segments separately and combining their signatures multiplicatively, rather than computing the full iterated integral from scratch.

Finally, for a given patient $i$, the set of extracted covariates over their follow-up period, is denoted as: \begin{equation*}
    \label{sig_covariates}
    \mathbb{S}_i = \Big( S^{(1)}, \dots, S^{(1,1)}, \dots, S^{(1,1,1)}, \dots, S^{(\bar{p}, \, \bar{p}, \, \bar{p})}  \Big)_i
. \end{equation*}

In order to ensure unicity of signature, the usual framework incorporates a monotonic component, especially time component. 

Thus, each patient's time series has been transformed into a set of covariates, providing a structured approach to handling sequential data. This transformation enables the extraction of meaningful temporal features, paving the way for their integration into a regression-based survival model, such as the Cox Proportional Hazards model, to assess patient-specific risk factors.

\subsection{Survival Analysis Modelling}

Our choice to illustrate the impact of textual data focused on the model of \cite{cox_og}, which accounts for censored patients, \textit{i.e.}, those for whom the event of interest $T$ (e.g., death, relapse) has not yet occurred. These observations still contribute to the likelihood estimation, helping to reduce bias and improve the robustness of predictions. The goal is to estimate the probability of a patient surviving beyond time $t$, noted as $\mathcal{S}(t\mid \mathbb{S}) := \mathbb{P} \big( T \geq t \mid \mathbb{S} \big)$ when knowing their covariates $\mathbb{S}$. This estimation relies on the key concept of instantaneous hazard rate $h$, which quantifies the infinitesimal probability of the event occurring at $t$ and is related to survival through the following equation: \begin{equation*}
    \label{ODE_risk_surv}
    \mathcal{S}(t\mid \mathbb{S}) = \exp\left( - \int^{t} h(s\mid \mathbb{S}) \, \mathrm{d}s\right)
. \end{equation*} Cox \cite{cox_og} proposed the generalized linear model: \begin{equation*}
    \label{cox_hazard_model}
    h(t\mid \mathbb{S}) \, = \, h_0(t) \cdot \exp\big( \mathbb{S} \cdot \boldsymbol{\beta} \big)
, \end{equation*} where $\boldsymbol{\beta} \in \mathbb{R}^{\bar{p}}$ is the vector of parameters to be estimated, and $h_0$ is the baseline hazard, common to all patients, as estimated by \cite{breslow1972}. We define $\eta := \mathbb{S} \cdot \boldsymbol{\beta}$, referred to as the risk score. Estimating $\boldsymbol{\beta}$ involves managing a substantial number of covariates. As mentioned earlier, this is due to the signature transform, which generates a high-dimensional feature space: for $p$ input channels and a truncation level $L$, the number of resulting signature coefficients scales as $O(p^L)$. Even after dimensionality reduction, the resulting covariate space remains large. To reduce the risk of overfitting and improve model stability, we apply the LASSO (\textit{Least Absolute Shrinkage and Selection Operator}) regularization to the Cox model, as originally introduced in \cite{Cox_LASSO}: \begin{equation}
    \label{lasso_pen}
    \widehat{\boldsymbol{\beta}}^{\ell_1} \in \underset{\boldsymbol{\beta}}{\operatorname{argmax}} \ \log \text{PL}(\boldsymbol{\beta}) - \lambda \| \boldsymbol{\beta} \|_1,
\end{equation} where $\text{PL}(\boldsymbol{\beta})$ is the partial likelihood defined in \cite{partial_likelihood}, and $\lambda > 0$ the regularization parameter. $ \|\boldsymbol{\beta}\|_1 $ is the $\ell_1$-norm of the parameters $\boldsymbol{\beta}$. The impact of LASSO regularization is twofold: it shrinks some coefficients towards zero, effectively removing less relevant covariates, and it selects only the most important predictors for survival, enhancing model stability. The objective function from \eqref{lasso_pen} to be minimized, with $\log$ applied, is explicitly formulated as: \begin{equation*}
    \label{objective_fct_lasso}
    \ell(\boldsymbol{\beta}) = \sum_{i: \delta_i = 1} \left[ \mathbb{S}_i \boldsymbol{\beta} - \log  \sum_{j \in \mathcal{R}_i} \exp\Big(\mathbb{S}_j \boldsymbol{\beta}\Big)  \right] - \lambda \sum_{k=1}^{\bar{p}} |\beta_k | 
. \end{equation*} This formulation provides an explicit likelihood function to be maximized algorithmically, with $\delta_i \in \{0,1\}$ determining whether an event (\textit{e.g.}, death) has been observed for patient $i$, associated with study duration $T_i$, and $\mathcal{R}_i$ represents the risk set \textit{i.e.}, set of individuals still at risk at time of $T_i$, that is $\mathcal{R}_i = \{ j : T_j \geq T_i \}$. Moreover, by enforcing sparsity, the LASSO-regularized Cox model significantly reduces the number of active covariates, leading to faster computational performance. This suggests that the model achieves a favorable balance between overfitting and underfitting, leveraging a compact and efficient representation of the risk factors while maintaining strong predictive power. Finally, the estimated risk score under LASSO regularization is obtained simply as the dot product: \begin{equation*}
    \label{risk_score_lasso}
    \widehat{\eta} = \mathbb{S} \cdot \widehat{\boldsymbol{\beta}}^{\ell_1}
. \end{equation*} Consequently, our methodology assigns each patient - characterized by a series of medical reports - an estimated risk score $\widehat{\eta}$, effectively capturing the temporal evolution of their clinical trajectory. This structured approach enables a comprehensive integration of narrative data into survival analysis.

\section{Experiments}
\subsection{Cohort}

This study complies with the General Data Protection Regulation (GDPR) and falls within the scope of scientific research conducted in the legitimate interest of cancer research, in accordance with Articles 6.1.f and 9.2.j of Regulation (EU) No. 2016/679. This project has been officially registered under the MR004 declaration (V3.2, 23/08/2021) at the L\'eon B\'erard Center, ensuring compliance with legal and ethical standards for processing health data. The data have been carefully anonymized and can only be used within the framework of this study. No patient were opposed to this study. To ensure the reliability of our data, we selected a study cohort consisting of patients hospitalized - at least once - at the L\'eon B\'erard Center from 2000 to 2024, with comprehensive follow-up throughout their medical care to ensure data completeness and accuracy. 

The dataset consists of a clean and structured text corpus containing 274,420 medical reports from 7,121 patients, among whom 4,983 are deceased and 2,138 are censored. Reports mainly include consultation reports (63\%) and hospital stay reports (32\%). Each patient has an average of 39 medical reports (sd 25), reflecting the longitudinal nature of the dataset. The dataset covers all types of cancer, allowing for broad applicability of the survival analysis. The most prevalent cancer types include breast cancer (25\%), gynecological cancers (9.7\%), gastrointestinal cancers (8.6\%), lung cancer (5.5\%), prostate cancer (7\%), endocrine tumors (4.6\%), among others. The median survival time in the cohort is 1,024 days (approximately 2 years and 10 months) and the study period spans from 1997 to 2020, covering the 5th to 95th percentile of diagnosis dates.

\subsection{Hyperparameter Search}

Our experimental setup is designed to ensure reproducibility, robustness, and realistic evaluation in a complex real-world clinical context. We emphasize that our NLP model, OncoBERT - fine-tuned on oncology-specific clinical notes - is used as is throughout the study without further task-specific adaptation, thereby reflecting practical deployment scenarios. The survival model was trained on a cohort of 3,560 patients (136,748 reports) and evaluated on a separate test set of 3,561 patients (137,672 reports), using a structured and stratified train-test split to preserve temporal and distributional consistency. To calibrate the model, we conducted a grid search for the LASSO regularization parameter $\lambda$ within the range [0.001, 10], using a fixed step size of 0.001. We selected the value that maximized the cross-validation concordance index (C-index) averaged over five independent validation folds, within the training dataset. Our careful hyperparameter tuning, combined with a large dataset and relevant baseline comparisons, supports the reliability of our results and confirms the model's ability to handle high-dimensional sequential text data.

\newpage

At this step, the longitudinal data were transformed using the signature method, effectively eliminating any temporal constraints that could arise when subdividing the dataset for cross-validation. The selection criterion aimed to maximize the C-index through five-fold cross-validation. Specifically, for each candidate value of $\lambda$, the model was trained on a partition of the training cohort and evaluated on held-out subsets - within the training set -, maintaining equal proportions across folds.

The optimal value was then chosen as the one yielding the highest mean C-index, promoting robust generalization and avoiding overfitting. This tuning process was crucial for balancing model sparsity and predictive accuracy.

\subsection{Performance Metrics}

The validation process involved random splitting, where the test set was divided into ten disjoint subsets. Model evaluation was then repeated independently on each of these subsets, allowing us to compute mean performance metrics along with their standard deviations. To ensure a comprehensive evaluation, we rely on well-established metrics: the concordance index (C-index), the time-dependent AUC and the Brier Score. These metrics collectively provide a robust assessment of the model's predictive performance by capturing complementary aspects of survival prediction accuracy. The C-index, a fundamental metric in survival analysis introduced by \cite{harrell1982evaluating}, measures the proportion of concordant pairs among all possible pairs. Specifically, if patient $j$ experiences the event before patient $i$, then the model should assign a higher risk score to $j$. It is expressed as followed: \begin{equation*}
    \label{cindex}
    \text{C-index} = \frac{\sum_{i,j} \mathbbm{1}_{\{T_j<T_i\}} \cdot \mathbbm{1}_{\{\widehat{\eta}_j>\widehat{\eta}_i\}} \cdot \delta_j}{\sum_{i,j} \mathbbm{1}_{\{T_j<T_i\}} \cdot \delta_j}
, \end{equation*} where $T_i$ and $T_j$ represent the observed survival times of patients $i$ and $j$, respectively. The estimated risk scores assigned by the model to these patients are denoted as $\widehat{\eta}_i$ and $\widehat{\eta}_j$. The indicator variable $\delta_j$ equals 1 if patient $j$ experienced the event, and 0 otherwise. A C-index of 1 characterizes a perfect model, while a C-index of 0.5 corresponds to random performance. A C-index above 0.7 is generally considered satisfactory. 

Its time-dependent counterpart, the td-AUC, is defined and then integrated over the relevant time interval (see \cite{kvamme2019coxtime}, \cite{polsterl2019evaluating}). It evaluates the model's ability to discriminate between patients who experience an event at time $t$ and those who survive beyond $t$. This allows for a more precise assessment of the model's predictive power at different time points. By incorporating dynamic survival probabilities, the td-AUC provides a temporal perspective on model performance and is particularly valuable in contexts where the ability to predict risk evolves over time. It is defined as:

\begin{equation*}
    \widehat{\text{AUC}}(t) = \frac{\sum_{i,j} \mathbbm{1}_{\{T_i > t\}} \mathbbm{1}_{\{T_j \leq t\}} \cdot \mathbbm{1}_{\{\widehat{\eta}_j > \widehat{\eta}_i\}} \cdot {\delta}_j(t)}{\sum_{i,j}\mathbbm{1}_{\{T_i > t\}} \mathbbm{1}_{\{T_j \leq t\}} \cdot {\delta}_j(t)}.
\end{equation*} 

Finally, the function also provides a single summary measure that refers to the mean of the $\text{AUC}(t)$ over the time range $[\tau_1, \tau_2]$:

\begin{equation*}
    \overline{\text{AUC}}(\tau_1, \tau_2) = 
    \frac{1}{\widehat{G}(\tau_1) - \widehat{G}(\tau_2)} 
    \int_{\tau_1}^{\tau_2} \widehat{\text{AUC}}(t) \, \mathrm{d}\widehat{G}(t),
\end{equation*} where $\widehat{G}(t)$ is the Kaplan-Meier estimator \cite{KaplanMeier1958} of the survival function. This accounts for censoring and provides a single summary measure of model performance across the specified time interval.

In recent years, growing attention has been paid to the calibration of survival models, not just their discriminative ability. While metrics such as the C-index remain central for assessing a model's ability to rank individuals by risk, they do not capture whether predicted survival probabilities are well-aligned with observed outcomes. As a result, proper evaluation now typically includes both discrimination and calibration metrics, to ensure that models are not only capable of ranking patients but also of assigning realistic survival probabilities. Calibration error is quantified by the Brier Score (BS) that evaluates the accuracy of survival predictions by assessing how close the estimated survival probability is to the actual survival status of each individual (also see \cite{kvamme2019coxtime}, \cite{polsterl2019evaluating}). With previous notations, and denoting $\mathcal{V}$ the set of individuals, the Brier Score can be expressed, at time $t$, as:
\begin{equation*}
    \mathrm{BS} (t) = \frac{1}{|\mathcal{V}|} 
    \sum_{{i} \in \mathcal{V}}  
    \left[ 
    \mathbbm{1}_{\{T_{i} \leq t\}}  
    \frac{\Big(0 - \widehat{\mathcal{S}} (t \mid \mathbb{X}_{i})\Big)^2}{\widehat{G}(T_{i})}\delta_{i}
    + 
    \mathbbm{1}_{\{T_{i} > t\}}  
    \frac{\Big(1 - \widehat{\mathcal{S}} (t \mid \mathbb{X}_{i})\Big)^2}{\widehat{G}(t)}
    \right].
\end{equation*}

Similar to the td-AUC, the Integrated Brier Score (IBS) provides a global average measure of calibration over a predefined time range $[\tau_1, \tau_2]$, defined as:

\begin{equation*}
    \mathrm{IBS}(\tau_1, \tau_2) = 
    \frac{1}{\tau_2 - \tau_1} 
    \int_{\tau_1}^{\tau_2} \mathrm{BS}(t) \, \mathrm{d}t.
\end{equation*}

A useful reference point for evaluating the BS (or the IBS) is the naive baseline, where the survival probability $\mathcal{S}(t)$ is set to a constant value of 0.5 for all individuals. In this case, the Brier Score simplifies to $\text{BS}(t) = 0.25$. Thus, a BS (or an IBS) below 0.25 is considered a good indicator of calibration.


\subsection{Experimental Results}

All our results are summarized in Table \ref{tab:results}. Our model achieves a mean C-index of 0.75 (sd 0.014) with a 95\% confidence interval [0.7419, 0.7596] (calculated by Jackknife method \cite{jackknife}), indicating good discriminative ability and suggesting the viability of our temporal approach. 

\begin{table}[h!]
    \centering
    \renewcommand{\arraystretch}{1.1}
    \begin{tabular}{|l|c|}
    \hline
    Metric & \textbf{Test Set} \\
    \hline
    Patients & 3,561 \\
    \hline
    Reports & 137 672 \\
    \hline
    C-index (mean) & 0.75 (sd 0.014) \\
    \hline
    CI$_{0.95}$ for C-index & [0.7419, 0.7596] \\
    \hline
    \multirow{2}{*}{Correlation \footnotesize{$\log$(Time) $\sim$ Risk}} & Pearson: -0.533 (sd 0.0359) \\
    & Spearman: -0.530 (sd 0.0459) \\
    \hline
    Mean td-AUC over 10 years & 0.794 (sd 0.029) \\
    \hline
    \multirow{3}{*}{Integrated Brier Score} & 3 years: 0.0532 (sd 0.0029) \\
    & 5 years: 0.1055 (sd 0.0062) \\
    & 10 years: 0.2183 (sd 0.0153) \\
    \hline    
    \end{tabular}
    \caption{Evaluation results for our pipeline.}
    \label{tab:results}
\end{table}

\begin{figure}
    \centering
    \includegraphics[width=\linewidth]{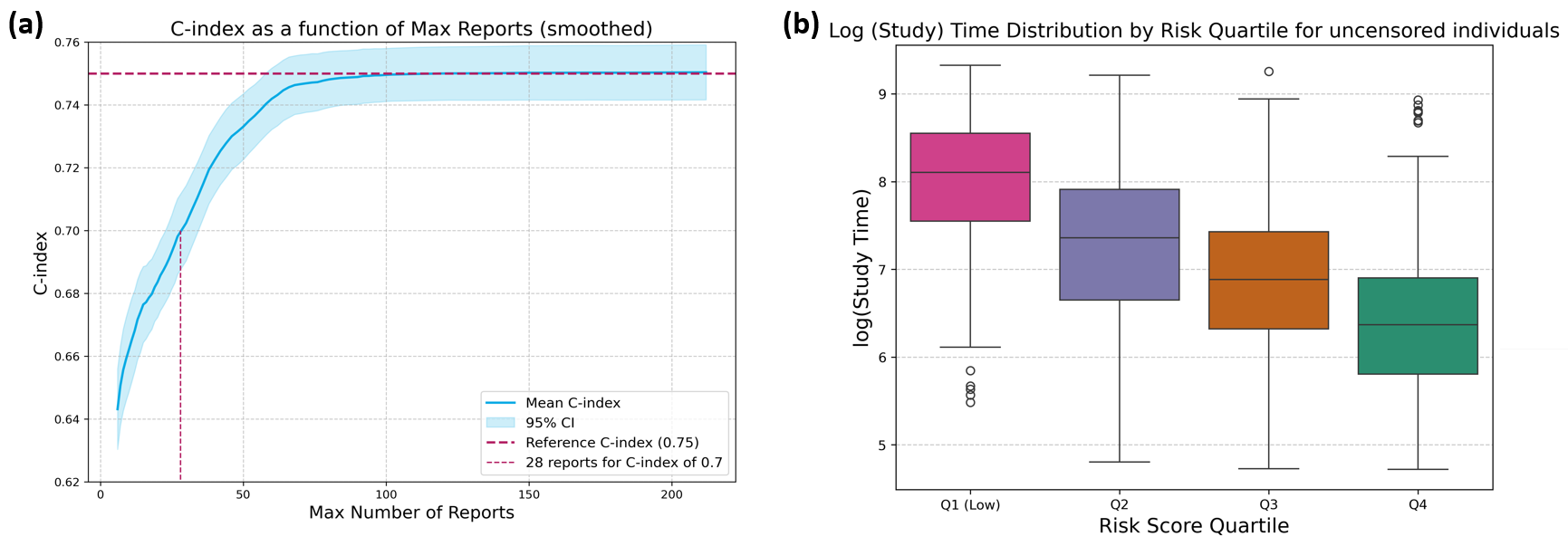}
    \caption{\textbf{(a) Test C-index progression as a function of the maximum number of known reports per patient; (b) Log-transformed study time distribution across risk quartiles.} 
    \textbf{(a)} The C-index starts at 0.63 with only two reports per patient and increases steadily, reaching 0.70 at 28 reports and converging to 0.75 beyond 100 reports. This highlights how access to a richer medical history improves prediction, especially with reports closer to the event. 
    \textbf{(b)} Boxplots illustrate the distribution of log-transformed survival times across predicted risk quartiles. Despite some natural overlap due to the complexity of survival data, a clear decreasing trend is observed: higher predicted risk scores align with shorter observed survival times. Both ANOVA and Kruskal-Wallis tests yield p-values below $10^{-5}$, confirming the statistical significance of this separation.}
    \label{fig:Cindex_and_quartiles}
\end{figure}

Figure \ref{fig:Cindex_and_quartiles} \textbf{(a)} illustrates the evolution of the C-index on the test set as a function of the number of known reports per patient. The evaluation starts with only two reports per patient, progressively incorporating one additional report at a time until the maximum available number is reached for each patient. The resulting monotonic increase in performance highlights the ability of the signature transform to effectively leverage sequential information, demonstrating its impact on improving the model’s predictive accuracy as more medical history is incorporated. This demonstrates that as the number of known time points increases, more information can be extracted, leading to a more accurate estimation of overall survival.

The mean td-AUC over 10 years is 0.794 (sd 0.029), and the IBS remains below 0.25 up to 10 years, reaching 0.0532 (sd 0.0029) at 3 years, suggesting that the model maintains good predictive accuracy while ensuring proper calibration. These results align with state-of-the-art model performance while being obtained in a complex real-world clinical setting, reinforcing the credibility of our approach.

The correlation between the logarithm of the time event for uncensored $\log (T)$ and the estimated risk score $\widehat{\eta}$ is significantly negative (Pearson: -0.533, Spearman: -0.530, both with p-value $< 10^{-5}$), demonstrating that higher estimated risk scores are associated with shorter survival times. This reinforces the model’s ability to capture meaningful risk stratification.


Figure \ref{fig:Cindex_and_quartiles} \textbf{(b)} further illustrates this relationship by displaying the distribution of log-transformed study durations across predicted risk quartiles. Despite some natural overlap due to the complexity of survival prediction, a clear trend emerges: Patients with shorter survival times tend to be classified into higher risk quartiles.

Additionally, the model effectively handles a large number of clinical reports, leveraging more than 100,000 medical documents for training and evaluation. The extensive dataset ensures robust performance assessment over numerous time points, highlighting the scalability and real-world applicability of our approach.


To assess its added value, we compared it to several naive baseline methods. First, we considered using only the last available clinical report for each patient, extracting its sentence embedding and feeding it directly into a Cox model; this resulted in a low C-index of 0.55. We then tested a simple average of all sentence embeddings per patient as input to the same model, which led to a slightly improved yet still unsatisfactory C-index of 0.57. Lastly, we evaluated the Cox-Time model \cite{kvamme2019coxtime}, which incorporates temporal features directly as covariates; it achieved a C-index of 0.63. These results demonstrate that simplistic representations or direct time encodings are insufficient to accurately capture the temporal complexity of patient trajectories. In contrast, our method, grounded in rigorous mathematical feature extraction, substantially improves performance, emphasizing the importance of modeling temporal dynamics with structured and principled approaches.

Our project relies on several specialized packages to ensure efficiency and accuracy throughout the pipeline. \texttt{SIF} \cite{arora2017asimple} implements the smooth inverse frequency method for computing sentence embeddings. The extraction of path signatures, a core component of our feature representation, is performed using \texttt{iisignature} \cite{iisignature}, which exploits Chen’s identity for efficient computation of iterated-integral signatures. For the survival model, we employ \texttt{skglm} \cite{skglm}, a high-performance package designed for generalized linear models, allowing efficient LASSO-penalized Cox regression. The estimation of survival functions, risk scores, and baseline hazards is handled through \texttt{lifelines} \cite{lifelines}, which provides a comprehensive framework for survival analysis. Finally, we use \texttt{sksurv} \cite{sksurv} to compute key evaluation metrics, such as the concordance index, time-dependent AUC, and Brier Score, supporting rigorous model assessment.

Furthermore, the use of compression, LASSO regularization, and Chen’s identity for signatures enables efficient training and inference, requiring only a few minutes once word embeddings are extracted, highlighting another unique advantage of our model.

\section{Conclusion}

Our model SigBERT highlights the potential of leveraging sequential textual data for survival analysis in oncology by introducing a structured and reproducible pipeline. Indeed, the judicious combination of a fine-tuned NLP model, signature transforms to capture the temporal progression of patient follow-ups, and a Cox model with LASSO regularization leads to consistent and promising results. These findings pave the way for extending the model to the entire patient database, enabling broader generalization of the approach.  

Among the limitations of our study, one key constraint lies in the necessity of compressing the original high-dimensional embeddings before applying the signature transform. Without this step, the number of coefficients to compute becomes prohibitively large, rendering the approach computationally intractable. While this compression might initially seem restrictive, it opens an interesting avenue of research into the sparsity and intrinsic geometry of the embedding space. It suggests that only a carefully selected subset or combination of embedding dimensions may be sufficient for effective risk estimation and survival prediction. 

Another important consideration is the dependency on OncoBERT, which, although specifically fine-tuned for our oncology dataset, can in principle be replaced with other natural language models. Furthermore, our model is currently best suited to data collected at the L\'eon B\'erard Center, where follow-up medical reports are systematically available. A more thorough assessment of the model's generalizability will require applying it to sequential textual data from patients in other institutions and clinical settings. To date, we have not benchmarked alternative language models on this specific survival task, which is partly due to the novelty of our approach (combining NLP embeddings with signature transforms and Cox survival modeling). As such, our pipeline represents a first step toward this direction, and future work will be needed to assess the impact of alternative embedding strategies within this framework.

A key avenue for improvement lies in integrating tabular and sequential data alongside textual embeddings. In future work, we plan to systematically compare different survival models to assess their relative performance. While we employed the Cox LASSO model due to its strong empirical results in our experiments, a broader benchmarking study is required. This will involve evaluating alternative approaches such as Random Survival Forests, Cox Neural Networks, and Cox Elastic Net, among others. Establishing a standardized comparison framework will allow us to identify the most robust and interpretable survival models for oncology risk estimation. Finally, by combining narrative and structured data - such as patient characteristics, disease features, and biological markers -, we aim to develop a multimodal survival model capable of providing more personalized and accurate oncology care.

\bigskip

\noindent \textbf{Acknowledgments.} This study was funded by SHAPE-Med@Lyon.

\bigskip

\noindent \textbf{Disclosure of Interests.} The other authors declare no potential conflicts of interest.

%
%
%

\bibliographystyle{splncs04}

\begin{thebibliography}{30}

\bibitem{arora2017asimple} Arora, S., Liang, Y., Ma, T.: A simple but tough-to-beat baseline for sentence embeddings. In: International Conference on Learning Representations (ICLR) (2017), published as a conference paper at ICLR 2017
\bibitem{skglm} Bertrand, Q., Klopfenstein, Q., Bannier, P.-A., Gidel, G., Massias, M.: Beyond L1: Faster and better sparse models with skglm. In: NeurIPS (2022)
\bibitem{CoxSig} Bleistein, L., Nguyen, V.T., Fermanian, A., Guilloux, A.: Dynamical survival analysis with controlled latent states (2024). arXiv:2401.17077 [stat.ML]. https://arxiv.org/abs/2401.17077
\bibitem{breslow1972} Breslow, N.E.: Contribution to the discussion of the paper by D. R. Cox. Journal of the Royal Statistical Society: Series B (Methodological) \textbf{34}(2), 216–217 (1972)
\bibitem{chen1954} Chen, K.T.: Iterated integrals and exponential homomorphisms. Proceedings of the London Mathematical Society \textbf{3}(4), 502–512 (1954)
\bibitem{Chevyrev2016} Chevyrev, I., Kormilitzin, A.: A primer on the signature method in machine learning (2025). arXiv:1603.03788 [stat.ML]. https://arxiv.org/abs/1603.03788
\bibitem{cox_og} Cox, D.R.: Regression models and life-tables. Journal of the Royal Statistical Society: Series B (Methodological) \textbf{34}(2), 187–202 (1972). https://doi.org/10.1111/j.2517-6161.1972.tb00899.x
\bibitem{partial_likelihood} Cox, D.R.: Partial likelihood. Biometrika \textbf{62}(2), 269–276 (1975). https://doi.org/10.1093/biomet/62.2.269
\bibitem{lifelines} Davidson-Pilon, C.: lifelines: Survival analysis in Python. Journal of Open Source Software \textbf{4}(40), 1317 (2019). https://doi.org/10.21105/joss.01317
\bibitem{Devaux2022} Devaux, A., Genuer, R., Peres, K., Proust-Lima, C.: Individual dynamic prediction of clinical endpoint from large dimensional longitudinal biomarker history: a landmark approach. BMC Medical Research Methodology \textbf{22}(1), 188 (2022). https://doi.org/10.1186/s12874-022-01660-3
\bibitem{jackknife} Efron, B.: The Jackknife, the Bootstrap and Other Resampling Plans. CBMS-NSF Regional Conference Series in Applied Mathematics, vol. 38. SIAM (1982). https://doi.org/10.1137/1.9781611970319
\bibitem{harrell1982evaluating} Harrell, F.E., Califf, R.M., Pryor, D.B., Lee, K.L., Rosati, R.A.: Evaluating the yield of medical tests. Journal of the American Medical Association \textbf{247}(18), 2543–2546 (1982)
\bibitem{Jee2024} Jee, J., Fong, C., Pichotta, K., et al.: Automated real-world data integration improves cancer outcome prediction. Nature \textbf{636}, 728–736 (2024). https://doi.org/10.1038/s41586-024-08167-5
\bibitem{KaplanMeier1958} Kaplan, E.L., Meier, P.: Nonparametric estimation from incomplete observations. Journal of the American Statistical Association \textbf{53}(282), 457–481 (1958). https://doi.org/10.2307/2281868
\bibitem{Katzman_2018} Katzman, J.L., Shaham, U., Cloninger, A., Bates, J., Jiang, T., Kluger, Y.: DeepSurv: personalized treatment recommender system using a Cox proportional hazards deep neural network. BMC Medical Research Methodology \textbf{18}(1) (2018). https://doi.org/10.1186/s12874-018-0482-1
\bibitem{kvamme2019coxtime} Kvamme, H., Borgan, {\O}., Scheel, I.: Time-to-event prediction with neural networks and Cox regression. Journal of Machine Learning Research \textbf{20}(116), 1–30 (2019). http://jmlr.org/papers/v20/18-424.html
\bibitem{dynamicdeephit2019} Lee, C., Yoon, J., van der Schaar, M.: Dynamic-DeepHit: A deep learning approach for dynamic survival analysis with competing risks based on longitudinal data. IEEE Transactions on Biomedical Engineering (2019). https://par.nsf.gov/servlets/purl/10099761
\bibitem{lee2018deephit} Lee, C., Zame, W.R., Yoon, J., van der Schaar, M.: DeepHit: A deep learning approach to survival analysis with competing risks. In: Proceedings of the Thirty-Second AAAI Conference on Artificial Intelligence (AAAI-18), pp. 2314–2321. AAAI Press (2018), Section: Discriminative Performance
\bibitem{lyons1998differential} Lyons, T.J.: Differential equations driven by rough signals. Revista Matemática Iberoamericana \textbf{14}(2), 215–310 (1998)
\bibitem{martin2020camembert} Martin, L., Muller, B., Ortiz Suárez, P.J., Dupont, Y., Romary, L., de la Clergerie, É., Seddah, D., Sagot, B.: CamemBERT: a tasty French language model. In: Proceedings of the 58th Annual Meeting of the Association for Computational Linguistics. Association for Computational Linguistics (2020). https://doi.org/10.18653/v1/2020.acl-main.645
\bibitem{dysurv2024} Mesinovic, M., Watkinson, P., Zhu, T.: DySurv: dynamic deep learning model for survival analysis with conditional variational inference (2024). arXiv:2310.18681 [cs.LG]. https://arxiv.org/abs/2310.18681
\bibitem{word2vec} Mikolov, T., Chen, K., Corrado, G., Dean, J.: Efficient estimation of word representations in vector space (2013). arXiv:1301.3781 [cs.CL]. https://arxiv.org/abs/1301.3781
\bibitem{sksurv} Pölsterl, S.: scikit-survival: A library for time-to-event analysis built on top of scikit-learn. Journal of Machine Learning Research \textbf{21}(212), 1–6 (2020). http://jmlr.org/papers/v21/20-729.html
\bibitem{survival_seq2seq} Pourjafari, E., Ziaei, N., Rezaei, M.R., Sameizadeh, A., Shafiee, M., Alavinia, M., Abolghasemian, M., Sajadi, N.: Survival Seq2Seq: A survival model based on sequence to sequence architecture (2022). arXiv:2204.04542 [cs.LG]. https://arxiv.org/abs/2204.04542
\bibitem{polsterl2019evaluating} Pölsterl, S.: Evaluating survival models (2019). https://k-d-w.org/blog/2019/05/evaluating-survival-models/
\bibitem{iisignature} Reizenstein, J., Graham, B.: The iisignature library: Efficient calculation of iterated-integral signatures and log signatures (2018). arXiv:1802.08252 [cs.DS]. https://arxiv.org/abs/1802.08252
\bibitem{Signorelli_2021} Signorelli, M., Spitali, P., Szigyarto, C.A., Tsonaka, R.: Penalized regression calibration: A method for the prediction of survival outcomes using complex longitudinal and high-dimensional data. Statistics in Medicine \textbf{40}(27), 6178–6196 (2021). https://doi.org/10.1002/sim.9178


\bibitem{Cox_LASSO} Tibshirani, R.: The Lasso method for variable selection in the Cox model. Statistics in Medicine \textbf{16}(4), 385–395 (1997). https://doi.org/10.1002/(SICI)1097-0258(19970228)16:4<385::AID-SIM380>3.0.CO;2-3
\bibitem{oncobert} Vienne, R., Filori, Q., Susplugas, V., Crochet, H., Verlingue, L.: Abstract 3475: Prediction of nausea or vomiting, and fatigue or malaise in cancer care. Cancer Research \textbf{84}, 3475–3475 (2024). https://doi.org/10.1158/1538-7445.AM2024-3475

\bibitem{bertsurv} Zhao, Y., Hong, Q., Zhang, X., Deng, Y., Wang, Y., Petzold, L.: BERTSurv: BERT-based survival models for predicting outcomes of trauma patients (2021). arXiv:2103.10928 [cs.AI]. https://arxiv.org/abs/2103.10928

\end{thebibliography}
%
%

\end{document}